# Twitter Sentiment Analysis

**By**

**Afroze Ibrahim Baqapuri (NUST-BEE-310)**

**A Project report submitted in fulfilment**

**of the requirement for the degree of**

**Bachelors in Electrical (Electronics) Engineering**

**Department of Electrical Engineering**

**School of Electrical Engineering & Computer Science**

**National University of Sciences & Technology**

**Islamabad, Pakistan**

**2012**

# CERTIFICATE

It is certified that the contents and form of thesis entitled **"Twitter Sentiment Analysis"** submitted by *Afroze Ibrahim Baqapuri* **(***NUST-BEE-310***)** have been found satisfactory for the requirement of the degree.

**Advisor: ______________________________**

**(Dr. Muhammad Usman Ilyas)**

**Co-Advisor: ______________________________**

**(Dr. Ali Mustafa Qamar)**

# DEDICATION

To Allah the Almighty

&

To my Parents and Faculty

# ACKNOWLEDGEMENTS

I am deeply thankful to my advisor and Co-Advisor, Dr. Muhammad Usman Ilyas, Dr. Ali Mustafa Qamar for helping me throughout the course in accomplishing my final project. Their guidance, support and motivation enabled me in achieving the objectives of the project.

# TABLE OF CONTENTS

# LIST OF FIGURES

# LIST OF TABLES

# ABSTRACT

This project addresses the problem of sentiment analysis in twitter; that is classifying tweets according to the sentiment expressed in them: positive, negative or neutral. Twitter is an online micro-blogging and social-networking platform which allows users to write short status updates of maximum length 140 characters. It is a rapidly expanding service with over 200 million registered users [24] - out of which 100 million are active users and half of them log on twitter on a daily basis - generating nearly 250 million tweets per day [20]. Due to this large amount of usage we hope to achieve a reflection of public sentiment by analysing the sentiments expressed in the tweets. Analysing the public sentiment is important for many applications such as firms trying to find out the response of their products in the market, predicting political elections and predicting socioeconomic phenomena like stock exchange. The aim of this project is to develop a functional classifier for accurate and automatic sentiment classification of an unknown tweet stream.

*Chapter 1*

# INTRODUCTION

## Motivation

We have chosen to work with twitter since we feel it is a better approximation of public sentiment as opposed to conventional internet articles and web blogs. The reason is that the amount of relevant data is much larger for twitter, as compared to traditional blogging sites. Moreover the response on twitter is more prompt and also more general (since the number of users who tweet is substantially more than those who write web blogs on a daily basis). Sentiment analysis of public is highly critical in macro-scale socioeconomic phenomena like predicting the stock market rate of a particular firm. This could be done by analysing overall public sentiment towards that firm with respect to time and using economics tools for finding the correlation between public sentiment and the firm's stock market value. Firms can also estimate how well their product is responding in the market, which areas of the market is it having a favourable response and in which a negative response (since twitter allows us to download stream of geo-tagged tweets for particular locations. If firms can get this information they can analyze the reasons behind geographically differentiated response, and so they can market their product in a more optimized manner by looking for appropriate solutions like creating suitable market segments. Predicting the results of popular political elections and polls is also an emerging application to sentiment analysis. One such study was conducted by Tumasjan et al. in Germany for predicting the outcome of federal elections in which concluded that twitter is a good reflection of offline sentiment [4].

## Domain Introduction

This project of analyzing sentiments of tweets comes under the domain of "Pattern Classification" and "Data Mining". Both of these terms are very closely related and intertwined, and they can be formally defined as the process of discovering "useful" patterns in large set of data, either automatically (unsupervised) or semi-automatically (supervised). The project would heavily rely on techniques of "Natural Language Processing" in extracting significant patterns and features from the large data set of tweets and on "Machine Learning" techniques for accurately classifying individual unlabelled data samples (tweets) according to whichever pattern model best describes them.

The features that can be used for modeling patterns and classification can be divided into two main groups: formal language based and informal blogging based. Language based features are those that deal with formal linguistics and include prior sentiment polarity of individual words and phrases, and parts of speech tagging of the sentence. Prior sentiment polarity means that some words and phrases have a natural innate tendency for expressing particular and specific sentiments in general. For example the word "excellent" has a strong positive connotation while the word "evil" possesses a strong negative connotation. So whenever a word with positive connotation is used in a sentence, chances are that the entire sentence would be expressing a positive sentiment. Parts of Speech tagging, on the other hand, is a syntactical approach to the problem. It means to automatically identify which part of speech each individual word of a sentence belongs to: noun, pronoun, adverb, adjective, verb, interjection, etc. Patterns can be extracted from analyzing the frequency distribution of these parts of speech (ether individually or collectively with some other part of speech) in a particular class of labeled tweets. Twitter based features are more informal and relate with how people express themselves on online social platforms and compress their sentiments in the limited space of 140 characters offered by twitter. They include twitter hashtags, retweets, word capitalization, word lengthening [13], question marks, presence of url in tweets, exclamation marks, internet emoticons and internet shorthand/slangs.

Classification techniques can also be divided into a two categories: Supervised vs. unsupervised and non-adaptive vs. adaptive/reinforcement techniques. Supervised approach is when we have pre-labeled data samples available and we use them to train our classifier. Training the classifier means to use the pre-labeled to extract features that best model the patterns and differences between each of the individual classes, and then classifying an unlabeled data sample according to whichever pattern best describes it. For example if we come up with a highly simplified model that neutral tweets contain 0.3 exclamation marks per tweet on average while sentiment-bearing tweets contain 0.8, and if the tweet we have to classify does contain 1 exclamation mark then (ignoring all other possible features) the tweet would be classified as subjective, since 1 exclamation mark is closer to the model of 0.8 exclamation marks. Unsupervised classification is when we do not have any labeled data for training. In addition to this adaptive classification techniques deal with feedback from the environment. In our case feedback from the environment can be in form of a human telling the classifier whether it has done a good or poor job in classifying a particular tweet and the classifier needs to learn from this feedback. There are two further types of adaptive techniques: Passive and active. Passive techniques are the ones which use the feedback only to learn about the environment (in this case this could mean improving our models for tweets belonging to each of the three classes) but not using this improved learning in our current classification algorithm, while the active approach continuously keeps changing its classification algorithm according to what it learns at real-time.

There are several metrics proposed for computing and comparing the results of our experiments. Some of the most popular metrics include: Precision, Recall, Accuracy, F1-measure, True rate and False alarm rate (each of these metrics is calculated individually for each class and then averaged for the overall classifier performance.) A typical confusion table for our problem is given below along with illustration of how to compute our required metric.

| | **Machine says yes** | **Machine says no** |
|---|---|---|
| **Human says yes** | **tp** | **fn** |
| **Human says no** | **fp** | **tn** |

**Table 1: A Typical 2x2 Confusion Matrix**

**Precision(P)** = $\frac{tp}{tp+fp}$ **Recall(R)** = $\frac{tp}{tp+fn}$ **Accuracy(A)** = $\frac{tp+tn}{tp+tn+f+fp+fn}$

**F1** = $\frac{2.P.R}{P+R}$ **True Rate(T)** = $\frac{tp}{tp+fn}$ **False-alarm Rate(F)** = $\frac{fp}{tp+fn}$

*Chapter 2*

# LITERATURE REVIEW

## Limitations of Prior Art

Sentiment analysis of in the domain of micro-blogging is a relatively new research topic so there is still a lot of room for further research in this area. Decent amount of related prior work has been done on sentiment analysis of user reviews [x], documents, web blogs/articles and general phrase level sentiment analysis [16]. These differ from twitter mainly because of the limit of 140 characters per tweet which forces the user to express opinion compressed in very short text. The best results reached in sentiment classification use supervised learning techniques such as Naive Bayes and Support Vector Machines, but the manual labelling required for the supervised approach is very expensive. Some work has been done on unsupervised (*e.g.,* [11] and [13]) and semi-supervised (*e.g.*, [3] and [10]) approaches, and there is a lot of room of improvement. Various researchers testing new features and classification techniques often just compare their results to base-line performance. There is a need of proper and formal comparisons between these results arrived through different features and classification techniques in order to select the best features and most efficient classification techniques for particular applications.

## Related Work

The bag-of-words model is one of the most widely used feature model for almost all text classification tasks due to its simplicity coupled with good performance. The model represents the text to be classified as a bag or collection of individual words with no link or dependence of one word with the other, i.e. it completely disregards grammar and order of words within the text. This model is also very popular in sentiment analysis and has been used by various researchers. The simplest way to incorporate this model in our classifier is by using unigrams as features. Generally speaking n-grams is a contiguous sequence of “n” words in our text, which is completely independent of any other words or grams in the text. So unigrams is just a collection of individual words in the text to be classified, and we assume that the probability of occurrence of one word will not be affected by the presence or absence of any other word in the text. This is a very simplifying assumption but it has been shown to provide rather good performance (for example in [7] and [2]). One simple way to use unigrams as features is to assign them with a certain prior polarity, and take the average of the overall polarity of the text, where the overall polarity of the text could simply be calculated by summing the prior polarities of individual unigrams. Prior polarity of the word would be positive if the word is generally used as an indication of positivity, for example the word “sweet”; while it would be negative if the word is generally associated with negative connotations, for example “evil”. There can also be degrees of polarity in the model, which means how much indicative is that word for that particular class. A word like “awesome” would probably have strong subjective polarity along with positivity, while the word “decent” would although have positive prior polarity but probably with weak subjectivity.

There are three ways of using prior polarity of words as features. The simpler un-supervised approach is to use publicly available online lexicons/dictionaries which map a word to its prior polarity. The Multi-Perspective-Question-Answering (MPQA) is an online resource with such a subjectivity lexicon which maps a total of 4,850 words according to whether they are “positive” or “negative” and whether they have “strong” or “weak” subjectivity [25]. The SentiWordNet 3.0 is another such resource which gives probability of each word belonging to positive, negative and neutral classes [15]. The second approach is to construct a custom prior polarity dictionary from our training data according to the occurrence of each word in each particular class. For example if a certain word is occurring more often in the positive labelled phrases in our training dataset (as compared to other classes) then we can calculate the probability of that word belonging to positive class to be higher than the probability of occurring in any other class. This approach has been shown to give better performance, since the prior polarity of words is more suited and fitted to a particular type of text and is not very general like in the former approach. However, the latter is a supervised approach because the training data has to be labelled in the appropriate classes before it is possible to calculate the relative occurrence of a word in each of the class. Kouloumpis et al. noted a decrease in performance by using the lexicon word features along with custom n-gram word features constructed from the training data, as opposed to when the n-grams were used alone [7].

The third approach is a middle ground between the above two approaches. In this approach we construct our own polarity lexicon but not necessarily from our training data, so we don’t need to have labelled training data. One way of doing this as proposed by Turney et al. is to calculate the prior semantic orientation (polarity) of a word or phrase by calculating it’s mutual information with the word “excellent” and subtracting the result with the mutual information of that word or phrase with the word “poor” [11]. They used the number of result hit counts from online search engines of a relevant query to compute the mutual information. The final formula they used is as follows:

$$\boldsymbol{Polarity}(\boldsymbol{phrase}) = \boldsymbol{log_2} \frac{\boldsymbol{hits}(\boldsymbol{phrase\ NEAR}\ \text{"excellent"}).\boldsymbol{hits}(\text{"poor"})}{\boldsymbol{hits}(\boldsymbol{phrase\ NEAR}\ \text{"poor"}).\boldsymbol{hits}(\text{"}\boldsymbol{excellent}\text{"})}$$

Where *hits(phrase NEAR “excellent”)* means the number documents returned by the search engine in which the phrase (whose polarity is to be calculated) and word “excellent” are co-occurring. While *hits(“excellent”)* means the number of documents retuned which contain the word “excellent”. Prabowo et al. have gone ahead with this idea and used a seed of 120 positive words and 120 negative to perform the internet searches [12]. So the overall semantic orientation of the word under consideration can be found by calculating the closeness of that word with each one of the seed words and taking and average of it. Another graphical way of calculating polarity of adjectives has been discussed by Hatzivassiloglou et al. [8]. The process involves first identifying all conjunctions of adjectives from the corpus and using a supervised algorithm to mark every pair of adjectives as belonging to the same semantic orientation or different. A graph is constructed in which the nodes are the adjectives and links indicate same or different semantic orientation. Finally a clustering algorithm is applied which divides the graph into two subsets such that nodes within a subset mainly contain links of same orientation and links between the two subsets mainly contain links of different orientation. One of the subsets would contain positive adjectives and the other would contain negative.

Many of the researchers in this field have used already constructed publicly available lexicons of sentiment bearing words (*e.g.*, [7], [12] and [16]) while many others have also explored building their own prior polarity lexicons (*e.g.*, [3], [10] and [11]).

The basic problem with the approach of prior polarity approach has been identified by Wilson et al. who distinguish between prior polarity and contextual polarity [16]. They say that the prior polarity of a word may in fact be different from the way the word has been used in the particular context. The paper presented the following phrase as an example:

*Philip Clapp, president of the National Environment Trust, sums up well the general thrust of the reaction of environmental movements: "There is no reason at all to believe that the polluters are suddenly going to become reasonable."*

In this example all of the four underlined words "trust", "well", "reason" and "reasonable" have positive polarities when observed without context to the phrase, but here they are not being used to express a positive sentiment. This concludes that even though generally speaking a word like "trust" may be used in positive sentences, but this doesn't rule out the chances of it appearing in non-positive sentences as well.

Henceforth prior polarities of individual words (whether the words generally carry positive or negative connotations) may alone not enough for the problem. The paper explores some other features which include grammar and syntactical relationships between words to make their classifier better at judging the contextual polarity of the phrase.

The task of twitter sentiment analysis can be most closely related to phrase-level sentiment analysis. A seminal paper on phrase level sentiment analysis was presented in 2005 by Wilson et al. [16] which identified a new approach to the problem by first classifying phrases according to subjectivity (polar) and objectivity (neutral) and then further classifying the subjective-classified phrases as either positive or negative. The paper noticed that many of the objective phrases used prior sentiment bearing words in them, which led to poor classification of especially objective phrases. It claims that if we use a simple classifier which assumes that the contextual polarity of the word is merely equal to its prior polarity gives a result of about 48%. The novel classification process proposed by this paper along with the list of ingenious features which include information about contextual polarity resulted in significant improvement in performance (in terms of accuracy) of the classification process. The results from this paper are presented in the table below:

| Features | Accuracy | Subjective F. | Objective F. |
|---|---|---|---|
| Word tokens | 73.6 | 55.7 | 81.2 |
| Words + prior polarity | 74.2 | 60.6 | 80.7 |
| 28 features | 75.9 | 63.6 | 82.1 |

Table 2: Step 1 results for Objective / Subjective Classification in [16]

| Features | Accuracy | Positive F. | Negative F. | Both F. | Objective F. |
|---|---|---|---|---|---|
| Word tokens | 61.7 | 61.2 | 73.1 | 14.6 | 37.7 |
| Word + prior | 63.0 | 61.6 | 75.5 | 14.6 | 40.7 |
| 10 features | 65.7 | 65.1 | 77.2 | 16.1 | 46.2 |

Table 3: Step 2 results for Polarity Classification in [16]

One way of alleviating the condition of independence and including partial context in our word models is to use bigrams and trigrams as well besides unigrams. Bigrams are collection of two contiguous words in a text, and similarly trigrams are collection of three contiguous words. So we could calculate the prior polarity of the bigram / trigram - or the prior probability of that bigram / trigram belonging to a certain class – instead of prior polarity of individual words. Many researchers have experimented with them with the general conclusion that if we have to use one of them alone unigrams perform the best, while unigrams along with bigrams may give better results with certain classifiers [2], [3]. However trigrams usually result in poor performance as reported by Pak et al. [3]. The reduction in performance by using trigrams is because there is a compromise between capturing more intricate patterns and word coverage as one goes to higher-numbered grams. Besides from this some researchers have tried to incorporate negation into the unigram word models. Pang et al. and Pakl et al. used a model in which the prior polarity of the word was reversed if there was a negation (like “not”, “no”, “don’t”, etc.) next to that word [5], [3]. In this way some contextual information is included in the word models.

Grammatical features (like “Parts of Speech Tagging” or POS tagging) are also commonly used in this domain. The concept is to tag each word of the tweet in terms of what part of speech it belongs to: noun, pronoun, verb, adjective, adverb, interjections, intensifiers etc. The concept is to detect patterns based on these POS and use them in the classification process. For example it has been reported that objective tweets contain more common nouns and third-person verbs than subjective tweets [3], so if a tweet to be classified has a proportionally large usage of common nouns and verbs in third person, that tweet would have a greater probability of being objective (according to this particular feature). Similarly subjective tweets contain more adverbs, adjectives and interjections [3]. These relationships are demonstrated in the figures below:

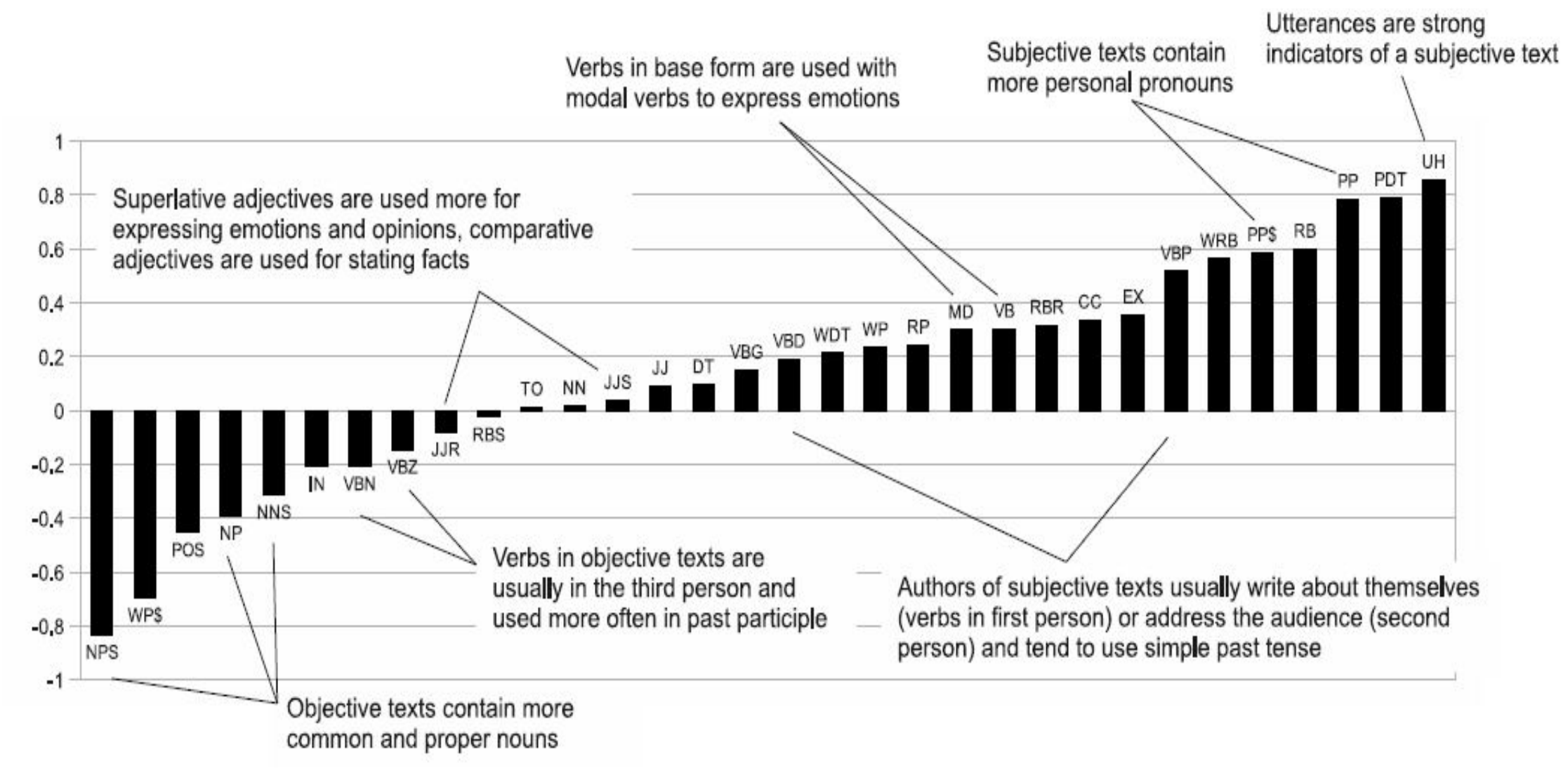

Figure 1: Using POS Tagging as features for objectivity/subjectivity classification

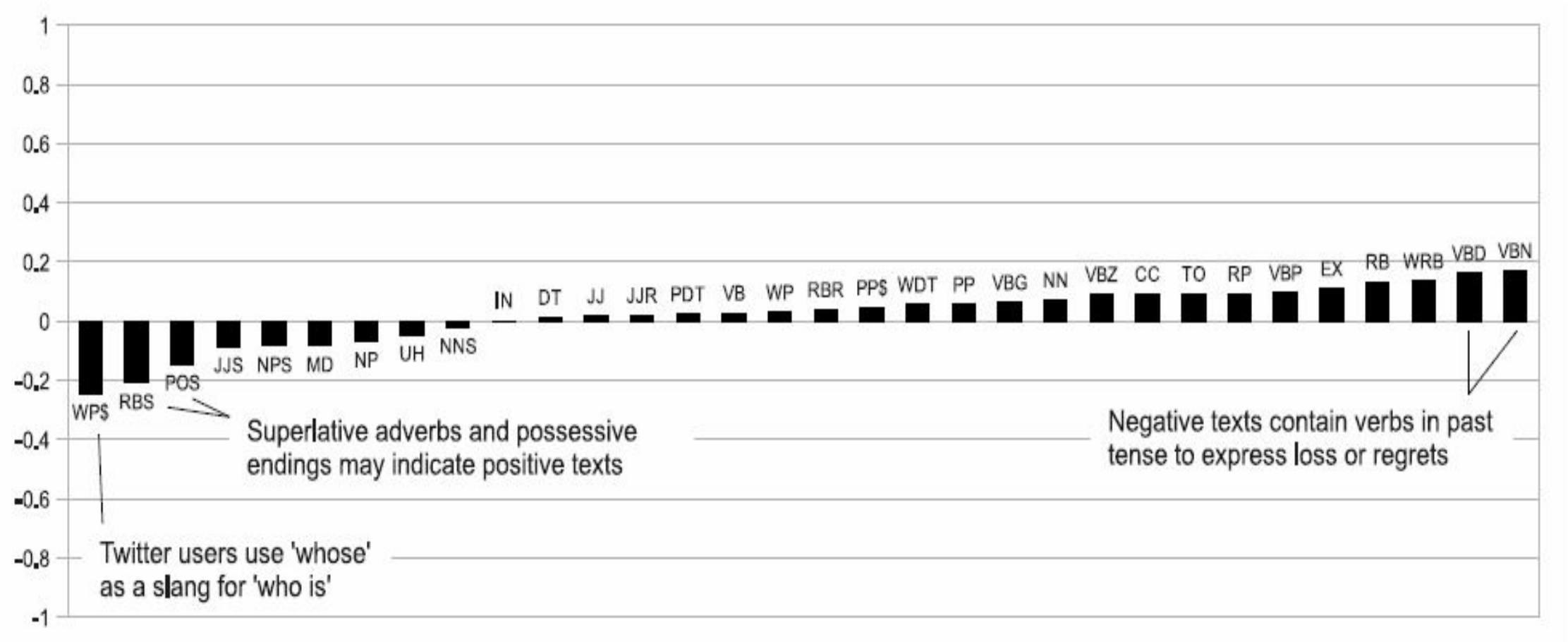

Figure 2: Using POS Tagging as features in positive/negative classification

However there is still conflict whether Parts-of-Speech are a useful feature for sentiment classification or not. Some researchers argue in favour of good POS features (*e.g.*, [10]) while others not recommending them (*e.g.*, [7]).

Besides from these much work has been done in exploring a class of features pertinent only to micro blogging domain. Presence of URL and number of capitalized words/alphabets in a tweet have been explored by Koulompis et al. [7] and Barbosa et al. [10]. Koulmpis also reports positive results for using emoticons and internet slang words as features. Brody et al. does study on word lengthening as a sign of subjectivity in a tweet [13]. The paper reports positive results for their study that the more number of cases a word has of lengthening, the more chance there of that word being a strong indication of subjectivity.

The most commonly used classification techniques are the Naive Bayes Classifier and State Vector Machines. Some researchers like Barbosa et al. publish better results for SVMs [10] while others like Pak et al. support Naive Bayes [3]. (1-9) and (2-6) also report good results for Maximum Entropy classifier.

It has been observed that having a larger training sample pays off to a certain degree, after which the accuracy of the classifier stays almost constant even if we keep adding more labelled tweets in the training data [10]. Barbosa et al. used tweets labelled by internet resources (*e.g.,* [28]), instead of labelling them by hand, for training the classifier. Although there is loss of accuracy of the labelled samples in doing so (which is modelled as increase in noise) but it has been observed that if the accuracy of training labels is greater than 50%, the more the labels, the higher the accuracy of the resulting classifier. So in this way if there are an extremely large number of tweets, the fact that our labels are noisy and inaccurate can be compensated for [10]. On the other hand Pak et al. and Go et al. [2] use presence of positive or negative emoticons to assign labels to the tweets [3]. Like in the above case they used large number of tweets to reduce effect of noise in their training data.

Some of the earliest work in this field classified text only as positive or negative, assuming that all the data provided is subjective (for example in [2] and [5]). While this is a good assumption for something like movie reviews but when analyzing tweets and blogs there is a lot of objective text we have to consider, so incorporating neutral class into the classification process is now becoming a norm. Some of the work which has included neutral class into their classification process includes [7], [10], [3] and [16].

There has also been very recent research of classifying tweets according to the mood expressed in them, which goes one step further. Bollen et al. explores this area and develops a technique to classify tweets into six distinct moods: tension, depression, anger, vigour, fatigue and confusion [9]. They use an extended version of Profile of Mood States (POMS): a widely accepted psychometric instrument. They generate a word dictionary and assign them weights corresponding to each of the six mood states, and then they represented each tweet as a vector corresponding to these six dimensions. However not much detail has been provided into how they built their customized lexicon and what technique did they use for classification.

*Chapter 3*

# FUNCTIONALITY AND DESIGN

The process of designing a functional classifier for sentiment analysis can be broken down into five basic categories. They are as follows:

I. Data Acquisition
II. Human Labelling
III. Feature Extraction
IV. Classification
V. TweetMood Web Application

## Data Acquisition:

Data in the form of raw tweets is acquired by using the python library "tweestream" which provides a package for simple twitter streaming API [26]. This API allows two modes of accessing tweets: SampleStream and FilterStream. SampleStream simply delivers a small, random sample of all the tweets streaming at a real time. FilterStream delivers tweet which match a certain criteria. It can filter the delivered tweets according to three criteria:

- Specific keyword(s) to track/search for in the tweets
- Specific Twitter user(s) according to their user-id's
- Tweets originating from specific location(s) (only for geo-tagged tweets).

A programmer can specify any single one of these filtering criteria or a multiple combination of these. But for our purpose we have no such restriction and will thus stick to the SampleStream mode.

Since we wanted to increase the generality of our data, we acquired it in portions at different points of time instead of acquiring all of it at one go. If we used the latter approach then the generality of the tweets might have been compromised since a significant portion of the tweets would be referring to some certain trending topic and would thus have more or less of the same general mood or sentiment. This phenomenon has been observed when we were going through our sample of acquired tweets. For example the sample acquired near Christmas and New Year's had a significant portion of tweets referring to these joyous events and were thus of a generally positive sentiment. Sampling our data in portions at different points in time would thus try to minimize this problem. Thus forth, we acquired data at four different points which would be 17$^{th}$ of December 2011, 29$^{th}$ of December 2011, 19$^{th}$ of January 2012 and 8$^{th}$ of February 2012.

A tweet acquired by this method has a lot of raw information in it which we may or may not find useful for our particular application. It comes in the form of the python "dictionary" data type with various key-value pairs. A list of some key-value pairs are given below:

- Whether a tweet has been favourited
- User ID
- Screen name of the user
- Original Text of the tweet
- Presence of hashtags
- Whether it is a re-tweet
- Language under which the twitter user has registered their account
- Geo-tag location of the tweet
- Date and time when the tweet was created

Since this is a lot of information we only filter out the information that we need and discard the rest. For our particular application we iterate through all the tweets in our sample and save the actual text content of the tweets in a separate file given that language of the twitter is user's account is specified to be English. The original text content of the tweet is given under the dictionary key "**text**" and the language of user's account is given under "**lang**".

Since human labelling is an expensive process we further filter out the tweets to be labelled so that we have the greatest amount of variation in tweets without the loss of generality. The filtering criteria applied are stated below:

- Remove Retweets (any tweet which contains the string "RT")
- Remove very short tweets (tweet with length less than 20 characters)
- Remove non-English tweets (by comparing the words of the tweets with a list of 2,000 common English words, tweets with less than 15% of content matching threshold are discarded)
- Remove similar tweets (by comparing every tweet with every other tweet, tweets with more than 90% of content matching with some other tweet is discarded)

After this filtering roughly 30% of tweets remain for human labelling on average per sample, which made a total of 10,173 tweets to be labelled.

## Human Labelling:

For the purpose of human labelling we made three copies of the tweets so that they can be labelled by four individual sources. This is done so that we can take average opinion of people on the sentiment of the tweet and in this way the noise and inaccuracies in labelling can be minimized. Generally speaking the more copies of labels we can get the better it is, but we have to keep the cost of labelling in our mind, hence we reached at the reasonable figure of three.

We labelled the tweets in four classes according to sentiments expressed/observed in the tweets: positive, negative, neutral/objective and ambiguous. We gave the following guidelines to our labellers to help them in the labelling process:

- **Positive**: If the entire tweet has a positive/happy/excited/joyful attitude or if something is mentioned with positive connotations. Also if more than one sentiment is expressed in the tweet but the positive sentiment is more dominant. Example: "*4 more years of being in shithole Australia then I move to the USA! :D*".
- **Negative**: If the entire tweet has a negative/sad/displeased attitude or if something is mentioned with negative connotations. Also if more than one sentiment is expressed in the tweet but the negative sentiment is more dominant. Example: "*I want an android now this iPhone is boring :S*".
- **Neutral/Objective**: If the creator of tweet expresses no personal sentiment/opinion in the tweet and merely transmits information. Advertisements of different products would be labelled under this category. Example: "*US House Speaker vows to stop Obama contraceptive rule... http://t.co/cyEWqKlE*".
- **Ambiguous**: If more than one sentiment is expressed in the tweet which are equally potent with no one particular sentiment standing out and becoming more obvious. Also if it is obvious that some personal opinion is being expressed here but due to lack of reference to context it is difficult/impossible to accurately decipher the sentiment expressed. Example: "*I kind of like heroes and don't like it at the same time...*". Finally if the context of the tweet is not apparent from the information available. Example: "*That's exactly how I feel about avengers haha*".
- **<Blank>**: Leave the tweet unlabelled if it belongs to some language other than English so that it is ignored in the training data.

Besides this labellers were instructed to keep personal biases out of labelling and make no assumptions, i.e. judge the tweet not from any past extra personal information and only from the information provided in the current individual tweet.

Once we had labels from four sources our next step was to combine opinions of three people to get an averaged opinion. The way we did this is through majority vote.

So for example if a particular tweet had to two labels in agreement, we would label the overall tweet as such. But if all three labels were different, we labelled the tweet as "unable to reach a majority vote". We arrived at the following statistics for each class after going through majority voting.

- Positive: 2543 tweets
- Negative: 1877 tweets
- Neutral: 4543 tweets
- Ambiguous: 451 tweets
- Unable to reach majority vote: 390 tweets
- Unlabelled non-English tweets: 369 tweets

So if we include only those tweets for which we have been able to achieve a positive, negative or neutral majority vote, we are left with 8963 tweets for our training set. Out of these 4543 are objective tweets and 4420 are subjective tweets (sum of positive and negative tweets).

We also calculated the human-human agreement for our tweet labelling task, results of which are as follows:

| | Human 1: Human 2 | Human 2: Human 3 | Human 1: Human 3 |
|---|---|---|---|
| Strict | 58.9% | 59.9% | 62.5% |
| Lenient | 65.1% | 67.1% | 73.0% |

**Table 4: Human-Human Agreement in Tweet Labelling**

In the above matrix the "strict" measure of agreement is where all the label assigned by both human beings should match exactly in all cases, while the "lenient" measure is in which if one person marked the tweet as "ambiguous" and the other marked it as something else, then this would not count as a disagreement. So in case of the "lenient" measure, the ambiguous class could map to any other class. So since the human-human agreement lies in the range of 60-70% (depending upon our definition of agreement), this shows us that sentiment classification is inherently a difficult task even for human beings. We will now look at another table presented by Kim et al. which shows human-human agreement in case labelling individual adjectives and verbs. [14]

| | Adjectives | Verbs |
|---|---|---|
| | Human 1: Human 2 | Human 1: Human 3 |
| Strict | 76.19% | 62.35% |
| Lenient | 88.96% | 85.06% |

**Table 5: Human- Human Agreement in Verbs / Adjectives Labelling [6]**

Over here the strict measure is when classification is between the three categories of positive, negative and neutral, while the lenient measure the positive and negative classes into one class, so now humans are only classifying between neutral and subjective classes. These results reiterate our initial claim that sentiment analysis is an inherently difficult task. These results are higher than our agreement results because in this case humans are being asked to label individual words which is an easier task than labelling entire tweets.

## Feature Extraction:

Now that we have arrived at our training set we need to extract useful features from it which can be used in the process of classification. But first we will discuss some text formatting techniques which will aid us in feature extraction:

- Tokenization: It is the process of breaking a stream of text up into words, symbols and other meaningful elements called "tokens". Tokens can be separated by whitespace characters and/or punctuation characters. It is done so that we can look at tokens as individual components that make up a tweet [19].
- Url's and user references (identified by tokens "http" and "@") are removed if we are interested in only analyzing the text of the tweet.
- Punctuation marks and digits/numerals may be removed if for example we wish to compare the tweet to a list of English words.
- Lowercase Conversion: Tweet may be normalized by converting it to lowercase which makes it's comparison with an English dictionary easier.
- Stemming: It is the text normalizing process of reducing a derived word to its root or stem [28]. For example a stemmer would reduce the phrases "stemmer", "stemmed", "stemming" to the root word "stem". Advantage of stemming is that it makes comparison between words simpler, as we do not need to deal with complex grammatical transformations of the word. In our case we employed the algorithm of "porter stemming" on both the tweets and the dictionary, whenever there was a need of comparison.
- Stop-words removal: Stop words are class of some extremely common words which hold no additional information when used in a text and are thus claimed to be useless [19]. Examples include "a", "an", "the", "he", "she", "by", "on", etc. It is sometimes convenient to remove these words because they hold no additional information since they are used almost equally in all classes of text, for example when computing prior-sentiment-polarity of words in a tweet according to their frequency of occurrence in different classes and using this polarity to calculate the average sentiment of the tweet over the set of words used in that tweet.

- Parts-of-Speech Tagging: POS-Tagging is the process of assigning a tag to each word in the sentence as to which grammatical part of speech that word belongs to, i.e. noun, verb, adjective, adverb, coordinating conjunction etc.

Now that we have discussed some of the text formatting techniques employed by us, we will move to the list of features that we have explored. As we will see below a feature is any variable which can help our classifier in differentiating between the different classes. There are two kinds of classification in our system (as will be discussed in detail in the next section), the objectivity / subjectivity classification and the positivity / negativity classification. As the name suggests the former is for differentiating between objective and subjective classes while the latter is for differentiating between positive and negative classes.

The list of features explored for objective / subjective classification is as below:

- Number of exclamation marks in a tweet
- Number of question marks in a tweet
- Presence of exclamation marks in a tweet
- Presence of question marks in a tweet
- Presence of url in a tweet
- Presence of emoticons in a tweet
- Unigram word models calculated using Naive Bayes
- Prior polarity of words through online lexicon MPQA
- Number of digits in a tweet
- Number of capitalized words in a tweet
- Number of capitalized characters in a tweet
- Number of punctuation marks / symbols in a tweet

- Ratio of non-dictionary words to the total number of words in the tweet
- Length of the tweet
- Number of adjectives in a tweet
- Number of comparative adjectives in a tweet
- Number of superlative adjectives in a tweet
- Number of base-form verbs in a tweet
- Number of past tense verbs in a tweet
- Number of present participle verbs in a tweet
- Number of past participle verbs in a tweet
- Number of 3$^{rd}$ person singular present verbs in a tweet
- Number of non-3$^{rd}$ person singular present verbs in a tweet
- Number of adverbs in a tweet
- Number of personal pronouns in a tweet
- Number of possessive pronouns in a tweet
- Number of singular proper noun in a tweet
- Number of plural proper noun in a tweet
- Number of cardinal numbers in a tweet
- Number of possessive endings in a tweet
- Number of wh-pronouns in a tweet
- Number of adjectives of all forms in a tweet
- Number of verbs of all forms in a tweet
- Number of nouns of all forms in a tweet
- Number of pronouns of all forms in a tweet

The list of features explored for positive / negative classification are given below:

- Overall emoticon score (where 1 is added to the score in case of positive emoticon, and 1 is subtracted in case of negative emoticon)

- Overall score from online polarity lexicon MPQA (where presence of strong positive word in the tweet increases the score by 1.0 and the presence of weak negative word would decrease the score by 0.5)
- Unigram word models calculated using Naive Bayes
- Number of total emoticons in the tweet
- Number of positive emoticons in a tweet
- Number of negative emoticons in a tweet
- Number of positive words from MPQA lexicon in tweet
- Number of negative words from MPQA lexicon in tweet
- Number of base-form verbs in a tweet
- Number of past tense verbs in a tweet
- Number of present participle verbs in a tweet
- Number of past participle verbs in a tweet
- Number of 3$^{rd}$ person singular present verbs in a tweet
- Number of non-3$^{rd}$ person singular present verbs in a tweet
- Number of plural nouns in a tweet
- Number of singular proper nouns in a tweet
- Number of cardinal numbers in a tweet
- Number of prepositions or coordinating conjunctions in a tweet
- Number of adverbs in a tweet
- Number of wh-adverbs in a tweet
- Number of verbs of all forms in a tweet

Next we will give mathematical reasoning of how we calculate the unigram word models using Naive Bayes. The basic concept is to calculate the probability of a word belonging to any of the possible classes from our training sample. Using mathematical formulae we will demonstrate an example of calculating probability of word belong to objective and subjective class. Similar steps would need to be taken for positive and negative classes as well.

We will start by calculating the probability of a word in our training data for belonging to a particular class:

$$P(word_1|obj) = \frac{count(word_1\ in\ obj\ class)}{count(total\ words\ in\ obj)}$$

We now state the Bayes' rule [19]. According to this rule, if we need to find the probability of whether a tweet is objective, we need to calculate the probability of tweet given the objective class and the prior probability of objective class. The term *P(tweet)* can be substituted with *P(tweet | obj) + P(tweet | subj).*

$$P(obj|tweet) = \frac{P(tweet|obj).P(obj)}{P(tweet)}$$

Now if we assume independence of the unigrams inside the tweet (i.e. the occurrence of a word in a tweet will not affect the probability of occurrence of any other word in the tweet) we can approximate the probability of tweet given the objective class to a mere product of the probability of all the words in the tweet belonging to objective class. Moreover, if we assume equal class sizes for both objective and subjective class we can ignore the prior probability of the objective class. Henceforth we are left with the following formula, in which there are two distinct terms and both of them are easily calculated through the formula mention above.

$$P(obj|tweet) = \frac{\prod_{i=1}^{N}\ [P(word_i|obj)}{\prod_{i=1}^{N}\ [P(word_i|obj) + \prod_{i=1}^{N}\ [P(word_i|subj)}$$

Now that we have the probability of objectivity given a particular tweet, we can easily calculate the probability of subjectivity given that same tweet by simply subtracting the earlier term from 1. This is because probabilities must always add to 1. So if we have information of *P(obj | tweet)* we automatically know *P(subj | tweet).*

$$P(subj|tweet) = 1 - P(obj|tweet)$$

Finally we calculate P(obj | tweet) for every tweet and use this term as a single feature in our objectivity / subjectivity classification.

There are two main potential problems with this approach. First being that if we include every unique word used in the data set then the list of words will be too large making the computation too expensive and time-consuming. To solve this we only include words which have been used at least 5 times in our data. This reduces the size of our dictionary for objective / subjective classification from 11,216 to 2,320. While for positive / negative classification unigram dictionary size is reduced from 6,502 to 1,235 words.

The second potential problem is if in our training set a particular word only appears in a certain class only and does not appear at all in the other class (for example if the word is misspelled only once). If we have such a scenario then our classifier will always classify a tweet to that particular class (regardless of any other features present in the tweet) just because of the presence of that single word. This is a very harsh approach and results in over-fitting. To avoid this we make use of the technique known as "Laplace Smoothing". We replace the formula for calculating the probability of a word belonging to a class with the following formula:

$$P(word_1|obj) = \frac{count(word_1\ in\ obj\ class) + x}{count(total\ words\ in\ obj) + x(total\ unique\ words\ in\ obj}$$

In this formula “x” is a constant factor called the smoothing factor, which we have arbitrarily selected to be 1. How this works is that even if the count of a word in a particular class is zero, the numerator still has a small value so the probability of a word belonging to some class will never be equal to zero. Instead if the probability would have been zero according to the earlier formula, it would be replace by a very small non-zero probability.

The final issue we have in feature selection is choosing the best features from a large number of features. Our ultimate aim is to achieve the greatest accuracy of our classifier while using least number of features. This is because adding new feature add to the dimensionality of our classification problem and thus add to the complexity of our classifier. This increase in complexity may not necessarily be linear and may even be quadratic so it is preferred to keep the features at a minimum low. Another issue we have with too many features is that our training data may be over-fit and it may confuse the classifier when doing classification on an unknown test set, so the accuracy of the classifier may even decrease. To solve this issue we select the most pertinent features by computing the information-gain of all the features under exploration and then selecting the features with highest information gain. We used WEKA machine learning tool for this task of feature selection [17].

We explored a total of 33 features for objectivity / subjectivity classification and used WEKA to calculate the information gain from each of these features. The resulting graph is shown below:

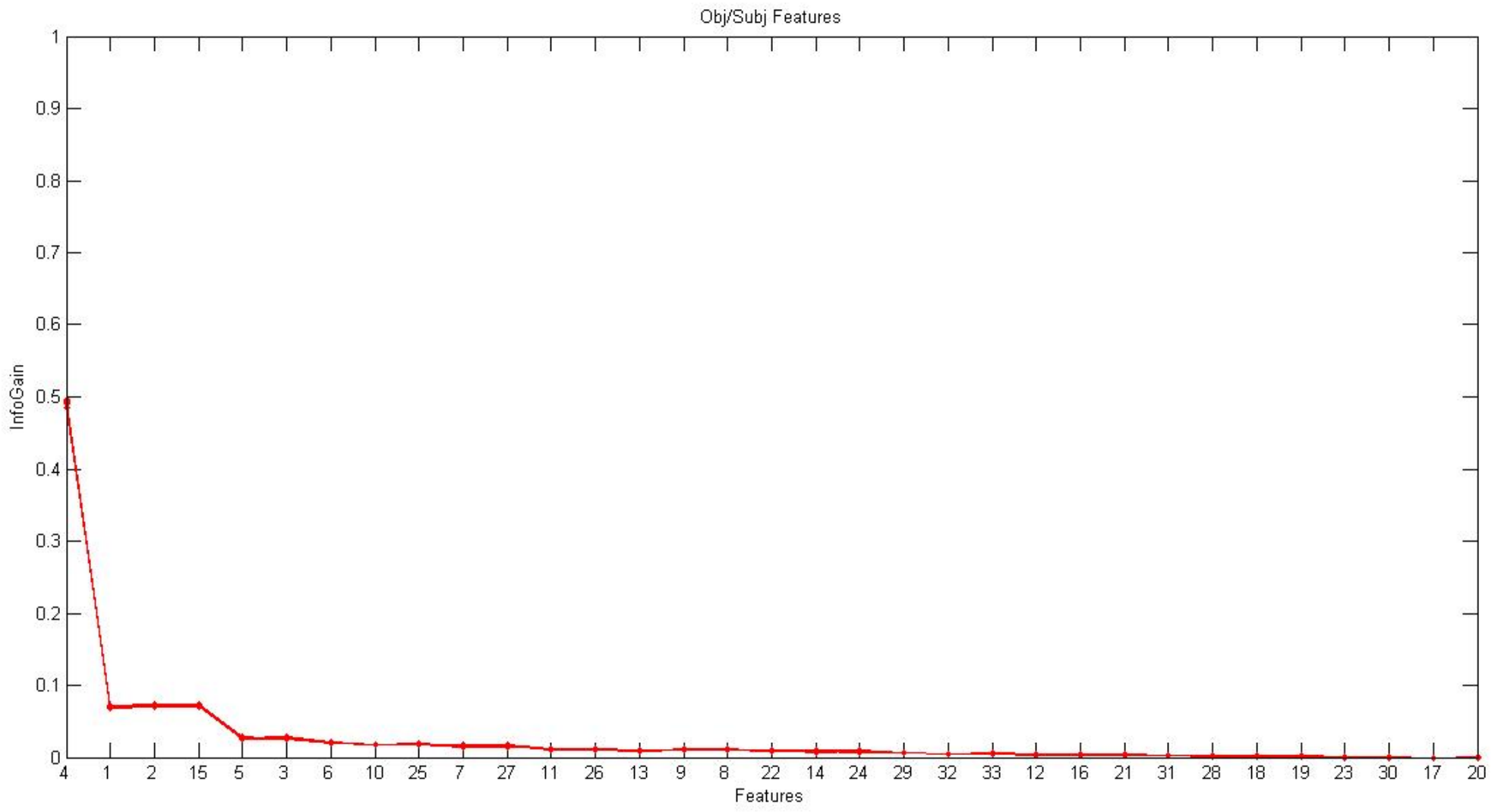

**Figure 3: Information Gain of Objectivity / Subjectivity Features**

This graph is basically the super-imposition of 10 different graphs, each one arrived through one fold out of the 10-fold cross validation we performed. Since we see that all the graphs are nicely overlapping so the results each fold are almost the same which shows us that the features we select will perform best in all the scenarios. We selected the best 5 features from this graph which are as follows:

1. Unigram word models (for prior probabilities of words belonging to objective / subjective classes)
2. Presence of URL in tweet
3. Presence of emoticons in tweet
4. Number of personal pronouns in tweet
5. Number of exclamation marks in tweet

Similarly we explored 22 features for positive / negative classification and used WEKA to calculate the information gain from each of these features. The resulting graph is shown below:

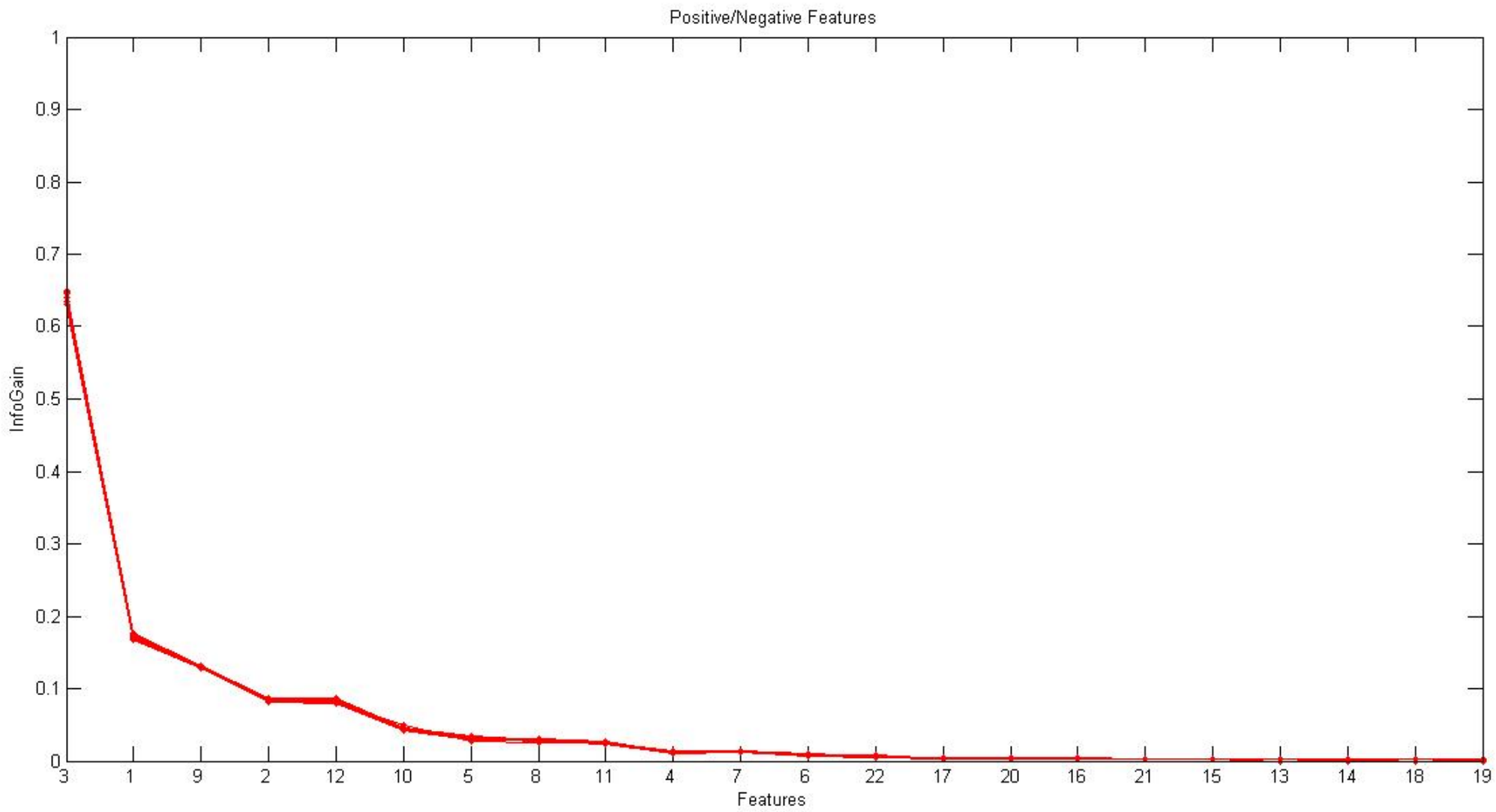

**Figure 4: Information Gain of Positive / Negative (Polarity) Features**

This graph is basically the super-imposition of 10 different graphs, each one arrived through one fold out of the 10-fold cross validation we performed. Since we see that all the graphs are nicely overlapping so the results each fold are almost the same which shows us that the features we select will perform best in all the scenarios. We selected the best 5 features out of which 2 were redundant features and we were left with only 3 features for our positive / negative classification which are as follows:

1. Unigram word models (for prior probabilities of words belonging to positive or negative classes)
2. Number of positive emoticons in tweet
3. Number of negative emoticons in tweet

The redundant features we chose ignore because they posed no extra information in presence of the above features are as follows:

- Emoticon score for the tweet
- MPQA score for the tweet

## Classification:

Pattern classification is the process through which data is divided into different classes according to some common patterns which are found in one class which differ to some degree with the patterns found in the other classes. The ultimate aim of our project is to design a classifier which accurately classifies tweets in the following four sentiment classes: positive, negative, neutral and ambiguous.

There can be two kinds of sentiment classifications in this area: contextual sentiment analysis and general sentiment analysis. Contextual sentiment analysis deals with classifying specific parts of a tweet according to the context provided, for example for the tweet "*4 more years of being in shithole Australia then I move to the USA :D*" a contextual sentiment classifier would identify Australia with negative sentiment and USA with a positive sentiment. On the other hand general sentiment analysis deals with the general sentiment of the entire text (tweet in this case) as a whole. Thus for the tweet mentioned earlier since there is an overall positive attitude, an accurate general sentiment classifier would identify it as positive. For our particular project we will only be dealing with the latter case, i.e. of general (overall) sentiment analysis of the tweet as a whole.

The classification approach generally followed in this domain is a two-step approach. First Objectivity Classification is done which deals with classifying a tweet or a phrase as either objective or subjective. After this we perform Polarity Classification (only on tweets classified as subjective by the objectivity classification) to determine whether the tweet is positive, negative or both (some researchers include the both category and some don't). This was presented by Wilson et al. and reports enhanced accuracy than a simple one-step approach [16].

We propose a novel approach which is slightly different from the approach proposed by Wilson et al. [16]. We propose that in first step each tweet should undergo two classifiers: the objectivity classifier and the polarity classifier. The former would try to classify a tweet between objective and subjective classes, while latter would do so between the positive and negative classes. We use the short-listed features for these classifications and use the Naive Bayes algorithm so that after the first step we have two numbers from 0 to 1 representing each tweet. One of these numbers is the probability of tweet belonging to objective class and the other number is probability of tweet belonging to positive class. Since we can easily calculate the two remaining probabilities of subjective and negative by simple subtraction by 1, we don't need those two probabilities.

So in the second step we would treat each of these two numbers as separate features for another classification, in which the feature size would be just 2. We use WEKA and apply the following Machine Learning algorithms for this second classification to arrive at the best result:

- K-Means Clustering
- Support Vector Machine
- Logistic Regression
- K Nearest Neighbours
- Naive Bayes
- Rule Based Classifiers

To better understand how this works we show a plot of actual test set from one of our cross-validations on the 2-dimensional space mentioned above:

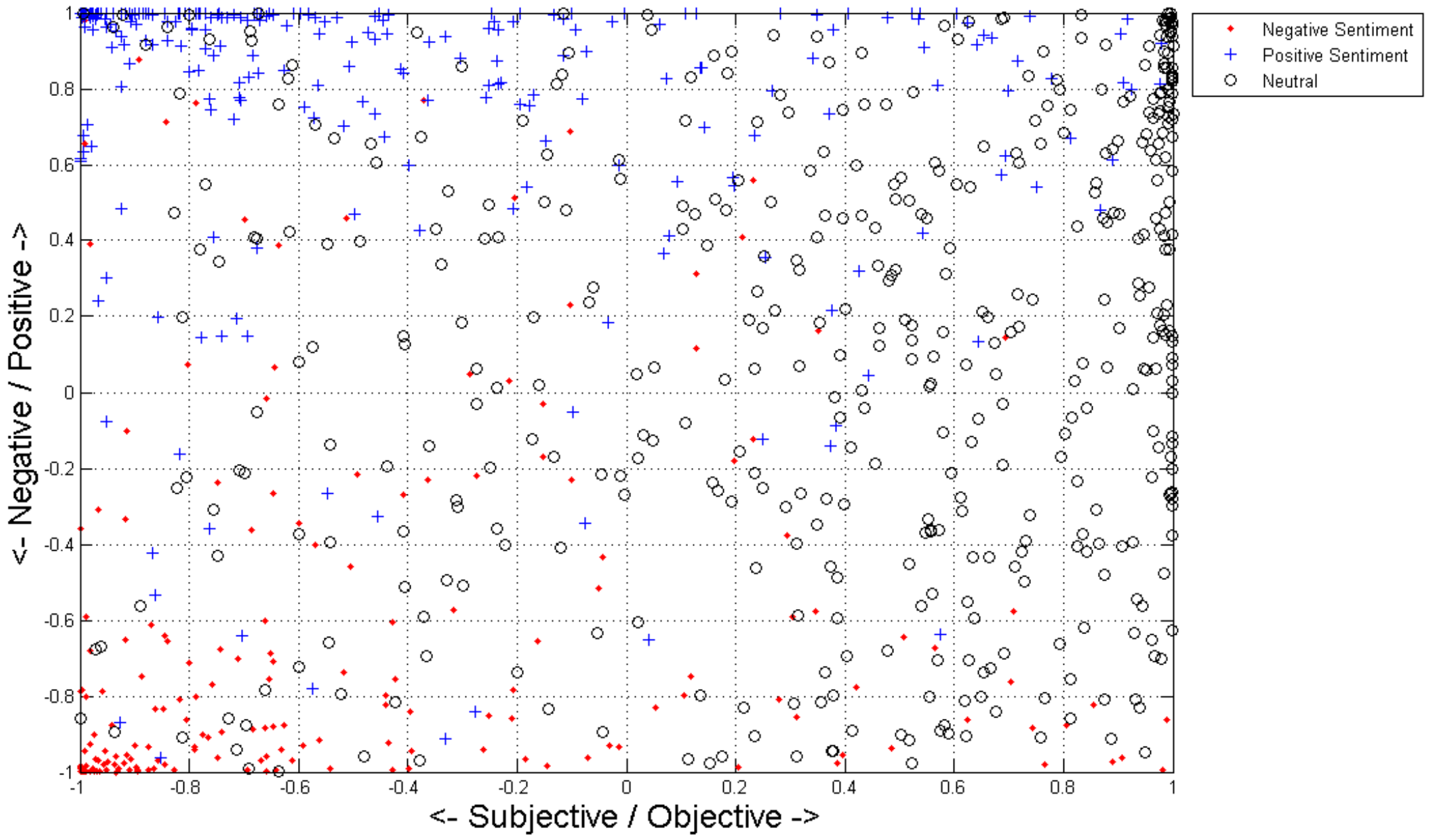

Figure 5: 2-d Scater Plot after Step 1

In this figure the labels are the actual ground truth and the distribution shows how the classified data points are actually scattered throughout the space. As we go right the tweet starts becoming increasingly objective and as we go up the tweet starts becoming more positive. The results for our classification approach are mentioned in the next section of this report.

## TweetMood Web Application:

We designed a web application which performed real-time sentiment analysis on Twitter on tweets that matched particular keywords provided by the user. For example if a user is interested in performing sentiment analysis on tweets which contain the word "Obama" he / she will enter that keyword and the web application will perform the appropriate sentiment analysis and display the results for the user.

The url of the website is www.tweet-mood-check.appspot.com and its logo is given below:

Figure 6: TweetMood web-application logo

The web application has been implemented using the Google App Engine service [21] because it can be used as a free web hosting service and it provides a layer of abstraction to the developer from the low level web operations so it is easier to learn. We implemented our algorithm in python and integrated it with GUI for our website using HTML and Javascript using the jinja2 template [23]. We used the Google Visualization Chart API for presenting our results in a graphical, easy-to-understand manner [22].

For acquiring tweets from Twitter we used the REST API in this case [27]. Twitter REST API provides access to tweets up to around 5 days in past according to the search query we specify. If we used the Twitter Streaming API and the user specified a keyword which is not very common in Twitter, the web application may have to wait for a long time to acquire enough tweets to display reasonable results. In contrast to this it is much simpler to acquire the tweets in a couple of simple URL calls to the Twitter REST API. One limitation of The REST API however is that one call can only give us a maximum of 100 results. Since we apply sentiment analysis on the past 1,000 tweets on any search query (given that there are that many tweets matching with the keyword available), so we have to basically call the API 10 times to get the required number of tweets. This is the basic source of processing delay in our web application.

We have three ways of performing sentiment analysis on our website and we will discuss each of them one by one:

- TweetScore
- TweetCompare
- TweetStats

**TweetScore:**

This feature calculates the popularity score of the keyword which is a number from 100 to -100. The more positive popularity score suggests that the keyword is highly positively popular on Twitter, while the more negative popularity score suggests that the keyword is highly negatively popular on Twitter. A popularity score close to 0 suggests that the keyword has either mixed opinions or is not a popular topic on Twitter. The popularity score is dependent on two ratios:

- Number of positive classified tweets / Number of negative classified tweets
- Number of tweets acquired / Time in past needed to explore the REST API

The first ratio suggests if the number of positive tweets is larger than negative tweets on a particular keyword, the keyword would have overall popular opinion and vice versa. The second ratio suggests that the lesser time in past we need to explore the REST API to get the 1,000 tweets means that the more number of people are talking about the keyword on Twitter, hence the keyword is popular on Twitter. However it gives no information about the positivity or negativity of the keyword and so higher the second ratio is, the more popularity score from the first ratio is shifted to the extreme ends (away from zero) may it be in positive or negative direction depends on whether there are more number of positive or negative tweets. Finally a maximum of 10 tweets are displayed for each class (positive, negative and neutral) so that the user develops confidence in our classifier.

**TweetCompare:**

This feature compares the popularity score of two or three different keywords and replies with which keyword is currently most popular on Twitter. This can have many interesting applications for example having our web application recommend users between movies, songs and products/brands.

**TweetStats:**

This feature is for long term sentiment analysis. We input a number of popular keywords on Twitter on which a backend operation runs after every hour, calculates the popularity score for the tweets generated on that keyword within an hour time frame and stores the results against every hour in a database. We can have a maximum of about 300 such keywords as per Google’s bandwidth requirements. So once we have a reasonable amount of data we can use it to plot graphs of popularity score against time and visualize the effect of change in popularity score with respect to certain events. Once we have collected enough data we can also use it to predict correlation with socio-economic phenomena like stock exchange rates and political elections. Work on this has been done before by Tumasjan et al. [4] and Bollen et al. [9].

*Chapter 4*

# IMPLEMENTATION AND RESULT DISCUSSION

We will first present our results for the objective / subjective and positive / negative classifications. These results act as the first step of our classification approach. We only use the short-listed features for both of these results. This means that for the objective / subjective classification we have 5 features and for positive / negative classification we have 3 features. For both of these results we use the Naïve Bayes classification algorithm, because that is the algorithm we are employing in our actual classification approach at the first step. Furthermore all the figures reported are the result of 10-fold cross validation. We take an average of each of the 10 values we get from the cross validation.

| Classes | True Positive | False Positive | Recall | Precision | F-measure |
|---|---|---|---|---|---|
| Objective | 0.73 | 0.26 | 0.74 | 0.73 | 0.73 |
| Subjective | 0.74 | 0.27 | 0.725 | 0.73 | 0.73 |
| Average | 0.73 | 0.27 | 0.73 | 0.73 | 0.73 |

**Table 6: Results from Objective / Subjective Classification**

| Classes | True Positive | False Positive | Recall | Precision | F-measure |
|---|---|---|---|---|---|
| Positive | 0.84 | 0.19 | 0.86 | 0.84 | 0.85 |
| Negative | 0.81 | 0.16 | 0.79 | 0.81 | 0.80 |
| Average | 0.83 | 0.18 | 0.83 | 0.83 | 0.83 |

**Table 7: Results from Polarity Classification (Positive / Negative)**

In addition to the above information, we make a condition while reporting the results of polarity classification (which differentiates between positive and negative classes) that only subjective labelled tweets are used to calculate these results. However, in case of final classification approach, any such condition is removed and basically both objectivity and polarity classifications are applied to all tweets regardless of whether they are labelled objective or subjective.

If we compare these results to those provided by Wilson et al. [16] (results are displayed in Table 2 and Table 3 of this report) we see that although the accuracy of neutral class falls from 82.1% to 73% if we use our classification instead of theirs. However, for all other classes we report significantly greater results. Although the results presented by Wilson et al. are not from Twitter data they are of phrase level sentiment analysis which is very close in concept to Twitter sentiment analysis.

Next we will compare our results with those presented by Go et al. [2]. The results presented by this paper are as follows:

| Features | Naive Bayes | Max Entropy | SVM |
|---|---|---|---|
| Unigram | 81.3% | 80.5% | 82.2% |
| Bigram | 81.6% | 79.1% | 78.8% |
| Unigram + Bigram | 82.7% | 83.0% | 81.6% |
| Unigram + POS | 79.9% | 79.9% | 81.9% |

**Table 8: Positive / Negative Classification Results presented by (1-9)**

If we compare these results to ours, we see that they are more or less similar. However, we arrive at comparable results with just 10 features and about 9,000 training data. In contrast to this, they used about 1.6 million noisy labels. Their labels were noisy in the sense that the tweets that contained positive emoticons were labelled as positive, while those with negative emoticons were labelled negative. The rest of the tweets (which did not contain any emoticon) were discarded from the data set. So in this way they hoped to achieve high results without human labelling but at the cost of using humongous large number amount of data set.

Next we will present our results for the complete classification. We note that the best results are reached through Support Vector Machine being applied at the second stage of the classification process. Hence the results below will only pertain to those of SVM. These results use a total of two features: P(objectivity | tweet) and P(positivity | tweet). But if we include all the features employed in step 1 of the classification process, we have a list of 8 shortlisted features (3 for polarity classification and 5 for objectivity classification). The following results are reported after conducting 10-fold cross validation:

| Classes | True Positive | False Positive | Recall | Precision | F-measure |
|---|---|---|---|---|---|
| Objective | 0.77 | 0.27 | 0.77 | 0.75 | 0.76 |
| Positive | 0.66 | 0.11 | 0.66 | 0.70 | 0.68 |
| Negative | 0.60 | 0.10 | 0.59 | 0.61 | 0.60 |
| Average | 0.70 | 0.19 | 0.703 | 0.703 | 0.703 |

**Table 9: Final Results using SVM at Step 2 and Naive Bayes at Step 1**

In comparison with these results, Koulompis et al. [7] reports average F-measure of 68%. However when they include another portion of their data into their classification process (which they call the HASH data), their average F-measure drops to 65%. In contrast to this we achieve average F-measure of more than 70% which shows better performance than either of these results. Moreover we make use of only 8 features and 9,000 labelled tweets, while their process involves about 15 features in total and more than 220,000 tweets in their training set. Our unigram word models are also simpler than theirs, because they incorporate negation into their word models. However like in the case of (1-9) their tweets are not labelled by humans, but rather undergo noisy labelling in two ways: labels acquired from positive and negative emoticons and hashtags.

Finally we conclude that our classification approach provides improvement in accuracy by using even the simplest features and small amount of data set. However there are still a number of things we would like to consider as future work which we mention in the next section.

*Chapter 5*

# CONCLUSION AND FUTURE RECOMMENDATIONS

The task of sentiment analysis, especially in the domain of micro-bloging, is still in the developing stage and far from complete. So we propose a couple of ideas which we feel are worth exploring in the future and may result in further improved performance.

Right now we have worked with only the very simplest unigram models; we can improve those models by adding extra information like closeness of the word with a negation word. We could specify a window prior to the word (a window could for example be of 2 or 3 words) under consideration and the effect of negation may be incorporated into the model if it lies within that window. The closer the negation word is to the unigram word whose prior polarity is to be calculated, the more it should affect the polarity. For example if the negation is right next to the word, it may simply reverse the polarity of that word and farther the negation is from the word the more minimized ifs effect should be.

Apart from this, we are currently only focusing on unigrams and the effect of bigrams and trigrams may be explored. As reported in the literature review section when bigrams are used along with unigrams this usually enhances performance.

However for bigrams and trigrams to be an effective feature we need a much more labeled data set than our meager 9,000 tweets.

Right now we are exploring Parts of Speech separate from the unigram models, we can try to incorporate POS information within our unigram models in future. So say instead of calculating a single probability for each word like *P(word | obj)* we could instead have multiple probabilities for each according to the Part of Speech the word belongs to. For example we may have *P(word | obj, verb), P(word | obj, noun) and P(word | obj, adjective).* Pang et al. [5] used a somewhat similar approach and claims that appending POS information for every unigram results in no significant change in performance (with Naive Bayes performing slightly better and SVM having a slight decrease in performance), while there is a significant decrease in accuracy if only adjective unigrams are used as features. However these results are for classification of reviews and may be verified for sentiment analysis on micro blogging websites like Twitter.

One more feature we that is worth exploring is whether the information about relative position of word in a tweet has any effect on the performance of the classifier. Although Pang et al. explored a similar feature and reported negative results, their results were based on reviews which are very different from tweets and they worked on an extremely simple model.

One potential problem with our research is that the sizes of the three classes are not equal. The objective class which contains 4,543 tweets is about twice the sizes of positive and negative classes which contain 2,543 and 1,877 tweets respectively. The problem with unequal classes is that the classifier tries to increase the overall accuracy of the system by increasing the accuracy of the majority class, even if that comes at the cost of decrease in accuracy of the minority classes. That is the very reason why we report significantly higher accuracies for objective class as opposed to positive or negative classes. To overcome this problem and have the classifier exhibit no bias towards any of the classes, it is necessary to label more data (tweets) so that all three of our classes are almost equal.

In this research we are focussing on general sentiment analysis. There is potential of work in the field of sentiment analysis with partially known context. For example we noticed that users generally use our website for specific types of keywords which can divided into a couple of distinct classes, namely: politics/politicians, celebrities, products/brands, sports/sportsmen, media/movies/music. So we can attempt to perform separate sentiment analysis on tweets that only belong to one of these classes (i.e. the training data would not be general but specific to one of these categories) and compare the results we get if we apply general sentiment analysis on it instead.

Last but not the least, we can attempt to model human confidence in our system. For example if we have 5 human labellers labelling each tweet, we can plot the tweet in the 2-dimensional objectivity / subjectivity and positivity / negativity plane while differentiating between tweets in which all 5 labels agree, only 4 agree, only 3 agree or no majority vote is reached. We could develop our custom cost function for coming up with optimized class boundaries such that highest weightage is given to those tweets in which all 5 labels agree and as the number of agreements start decreasing, so do the weights assigned. In this way the effects of human confidence can be visualized in sentiment analysis.